\documentclass[11pt]{article}

\usepackage[utf8]{inputenc}
\usepackage[T1]{fontenc}
\usepackage{xcolor}
\usepackage{graphicx}
\usepackage{subfig}
\usepackage{wrapfig}
\usepackage{hyperref}
\usepackage{url}
\usepackage{booktabs} 
\usepackage{abstract}
\usepackage{multicol}
\usepackage{fancyhdr}
\usepackage{amsmath}
\usepackage{amsfonts}
\usepackage{amsthm}
\usepackage{amssymb}
\usepackage{nicefrac}
\usepackage{microtype}
\usepackage{paralist}
\usepackage[font=small,labelfont=bf]{caption}
\usepackage[ruled,norelsize]{algorithm2e}
\usepackage[noend]{algorithmic}
\usepackage{calc}
\usepackage{tikz}
\usepackage{bbm}
\usepackage{enumitem}
\usepackage{scalerel}

\newcommand{\R}{\mathbb{R}}
\newcommand{\id}{\mathrm{I}}

\newcommand{\Normal}{\mathcal{N}}

\newcommand{\bmat}{\begin{pmatrix}}
\newcommand{\emat}{\end{pmatrix}}

\title{\vspace*{-1.5cm}Challenges of Convex Quadratic\\Bi-objective Benchmark Problems}

\fancypagestyle{firststyle}
{
   \fancyhf{}
   \fancyfoot{}
}

\author{
	Tobias Glasmachers\\
	Institute for Neural Computation\\
	Ruhr-University Bochum, Germany\\
	\texttt{tobias.glasmachers@ini.rub.de}
}

\date{}

\begin{document}

\maketitle

\setlength{\absleftindent}{2cm}
\setlength{\absrightindent}{2cm}

\begin{abstract}
\normalsize
Convex quadratic objective functions are an important base case in
state-of-the-art benchmark collections for single-objective
optimization on continuous domains. Although often considered rather
simple, they represent the highly relevant challenges of
non-separability and ill-conditioning. In the multi-objective case,
quadratic benchmark problems are under-represented. In this paper we
analyze the specific challenges that can be posed by quadratic functions
in the bi-objective case. Our construction yields a full factorial
design of 54 different problem classes. We perform experiments with
well-established algorithms to demonstrate the insights that can be
supported by this function class. We find huge performance differences,
which can be clearly attributed to two root causes: non-separability and
alignment of the Pareto set with the coordinate system.
\end{abstract}

\begin{multicols}{2}

\section{Introduction}

Empirical comparisons play a major role in evolutionary computation.
They provide valuable insights about many practical questions that
cannot be answered through theoretical analysis. Benchmarking studies
can serve at least two very different purposes. The first is to reveal
strengths and limitations of algorithms on previously identified
challenges. To this end, benchmark problems are designed very
specifically to pose one challenge, and usually only that very
challenge, which allows to analyze different aspects in isolation. The
second purpose is to test performance in application domains, often with
functions that resemble real-world problems in a computationally
tractable manner. In this paper we stick to the first paradigm.

Lists of properties making up good benchmark problems can be long and to
some extent contradictory. Here we focus particularly on multi-objective
(MO) optimization in continuous search spaces. For standard MO
definitions and nomenclature we refer to the literature \cite{zdt}. Our
desiderata or design goals for benchmark problems are listed in the
following:
\begin{itemize}
\item[G1]
	scalable to any search space dimension $d$ (number of variables),
\item[G2]
	instantiable, i.e., clear how to construct (fixed or random)
	instances of the class of problems,
\item[G3]
	known optimum, i.e., known Pareto front and Pareto set,
\item[G4]
	known optimal $\mu$-distribution, i.e., optimal placement of $\mu$
	points maximizing a performance indicator like dominated
	hypervolume,
\item[G5]
	known challenges posed by each single objective, like
	non-separability, ill-conditioning, and multi-modality,
\item[G6]
	known challenges of the multi-objective problem, like a convex or
	concave Pareto front,
\item[G7]
	non-trivial and non-degenerate Pareto set.
\end{itemize}
There are good reasons to extend this list further, for example with the
criteria listed in the excellent review \cite{wfg-review}. However, in
this paper we do not aim to provide a benchmark collection that covers
all possible needs. We pursue two goals: we explore the utility of
convex quadratic function for the construction of multi-objective
benchmark problems, and doing so, we aim to cover the above list of
design goals, which is somewhat tailored to convex quadratic problems.

There exist several collections of benchmark problems that are used
frequently to assess properties of evolutionary multi-objective
optimization algorithms. We limit ourselves to collections of generic
multi-objective problems, without a particular focus.
%, e.g., on many-objective problems with $m \gg 3$ objectives.
None of the existing collections fulfills all of the above goals. Here
we highlight rather different examples of benchmark suites following
clear design principles:
\begin{itemize}
\item
	%% TODO: explain the ZDT and DTLZ constructions in more detail!
	%%
	The scalable ZDT \cite{zdt} and DTLZ \cite{dtlz} problems are
	constructed around a known Pareto set and a corresponding front. At
	least in the bi-objective case, optimal or near-optimal
	$\mu$-distributions are known \cite{auger:2009,glasmachers2014b}.
	Knowing Pareto set, Pareto front and optimal $\mu$-distribution
	is sufficient to assess the convergence behavior with respect to
	basically all convergence and spread indicators, and to define
	target values for algorithm comparisons as done in the bi-objective
	BBOB suite.
	There is only one instance per problem and it is unclear how to
	generalize them to whole classes, so that variants could be
	constructed. Several MO-specific challenges are known by
	construction, but the characteristics of the individual functions
	are not considered.
	The Pareto set is located on an edge of the feasible region, and it
	is hence aligned with the coordinate system, which is arguably an
	unrealistically simple case (\cite{mocmaes} addresses this issue).
	These problems fulfill goals G1, G3, G4, and G6. Some of the
	shortcomings of these benchmarks sets were addressed by the walking
	fish group benchmarks~\cite{wfg}, however, with a strong focus on
	various sub-goal of~G6.
	Similar properties apply to the LSMOP benchmarks \cite{LSMOP}, which
	are largely based on DTLZ and WFG.
\item
	The bi-objective BBOB problem suite \cite{tusar2016coco} consists
	of 55 problems constructed through pairing of single-objective (SO)
	problems drawn from 10 function classes. The functions inherit
	scalability, variability, and known challenges from the well-known
	single-objective BBOB problems, however, the resulting Pareto sets
	and fronts are not known analytically, and it is not clear how the
	SO challenges translate into MO challenges. Bi-objective BBOB
	fulfills goals G1, G2, G5, and G7.
\item
	The very recent work of \cite{toure2019} is closest to our
	approach. It goes into great detail in investigating theoretical
	properties of bi-objective functions. However, our construction of
	problems with a simple Pareto set is significantly more general.
\end{itemize}
We find that different construction methods result in different goals
being reached. The first two goals are rather easy to cover, and so is
G4 in the bi-objective case, provided that G3 is fulfilled. However,
goal G3 is hard to achieve when starting from single-objective problems,
which is an attractive approach since it can potentially leverage the
large body of work done in the single-objective domain. Also, there is a
clear tension between goals G5 and G6: while the BBOB-style construction
of problems from single objectives does not yields analytical knowledge
of the Pareto front, the ZDT/DTLZ-style construction does not provide
much control over the objectives in isolation. In this paper we aim to
resolve this conflict by providing a construction method that fulfills
all of the above goals simultaneously.

\section{Quadratic Bi-Objective Problems}

Convex quadratic objective functions on $\R^d$ of the form $x \mapsto
\frac12 x^T H x + g x + c$ with positive definite Hessian matrix $H$ are
of significant relevance, simply because every twice continuously
differentiable function is locally approximated by its second order
Taylor expansion, and under the mild regularity condition of a strictly
positive definite Hessian in a (local) optimum it is of the above form.

\subsection{The Single-Objective Case}

For the above reason, in the single-objective case, fast convergence on
a large class of functions is ensured by fast convergence on convex
quadratic problems. The Covariance Matrix Adaptation (CMA) mechanism
\cite{hansen:2001} found in most modern evolution strategies suits this
challenge well.

\begin{figure*}
\begin{center}
	\includegraphics[width=0.24\textwidth]{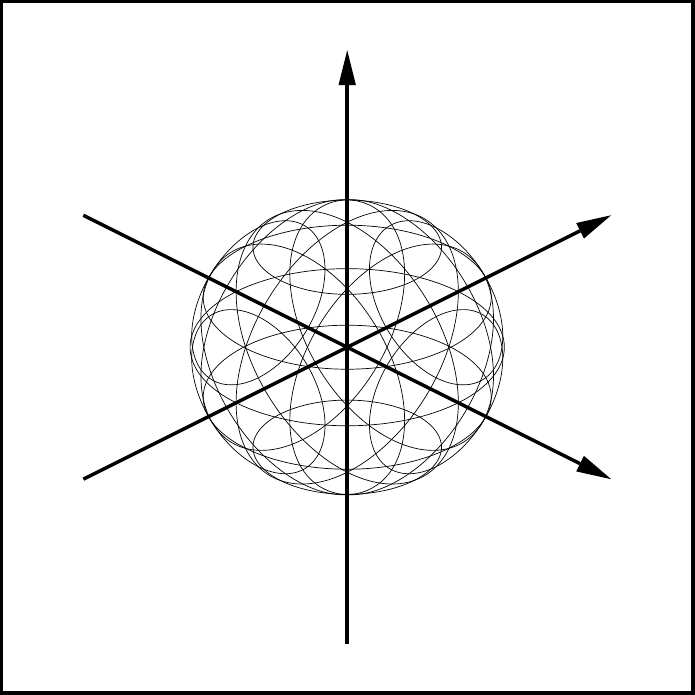}
	\includegraphics[width=0.24\textwidth]{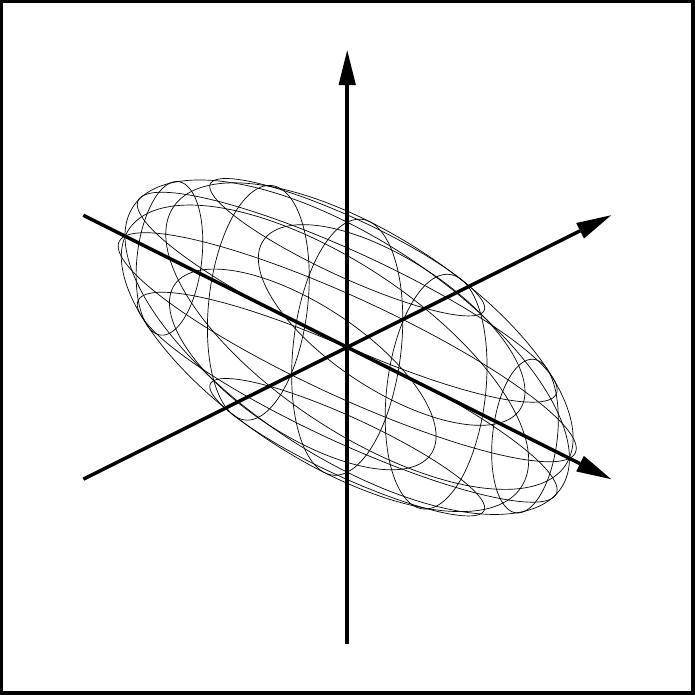}
	\includegraphics[width=0.24\textwidth]{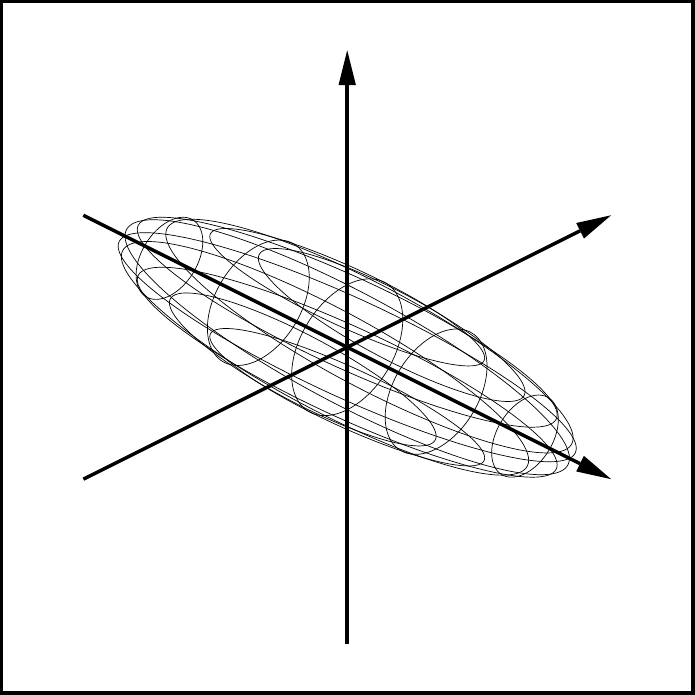}
	\includegraphics[width=0.24\textwidth]{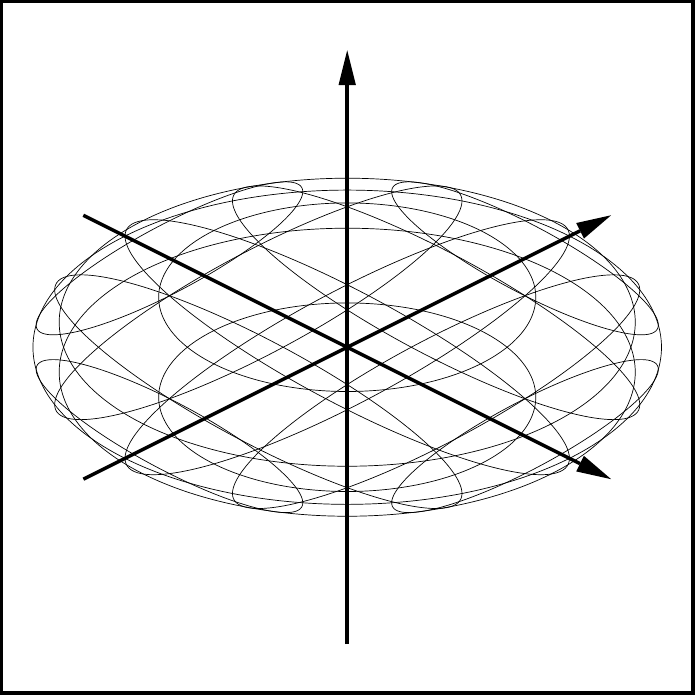}
	\\[2mm]
	\includegraphics[width=0.24\textwidth]{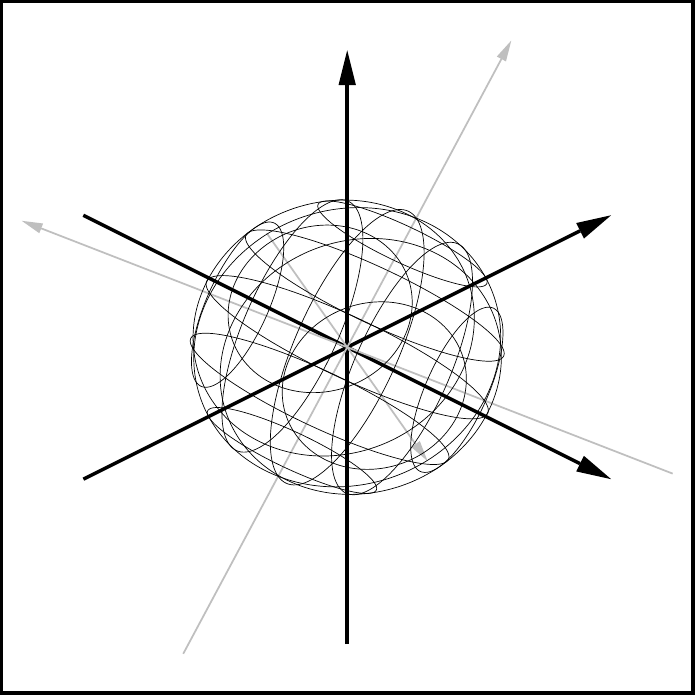}
	\includegraphics[width=0.24\textwidth]{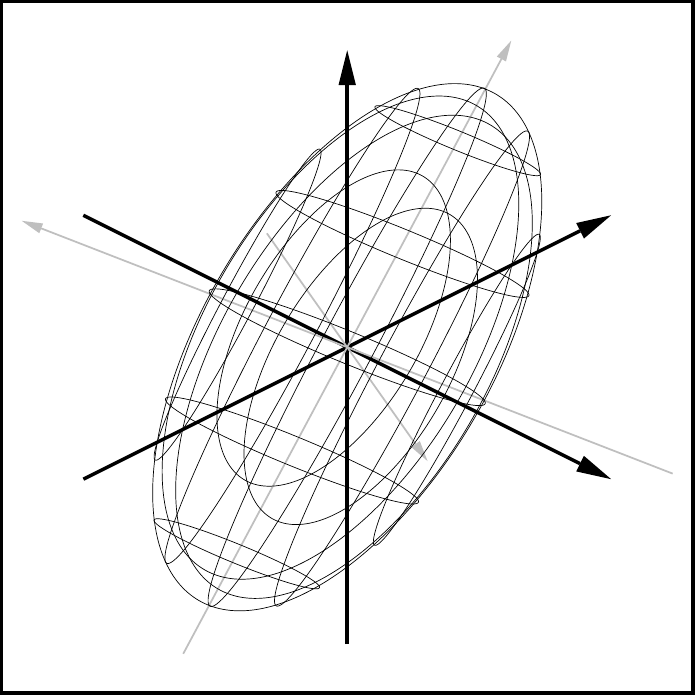}
	\includegraphics[width=0.24\textwidth]{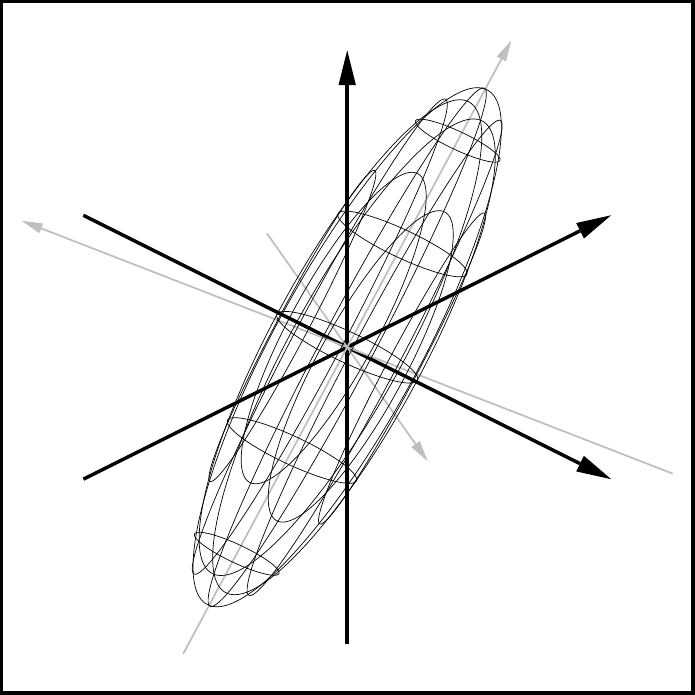}
	\includegraphics[width=0.24\textwidth]{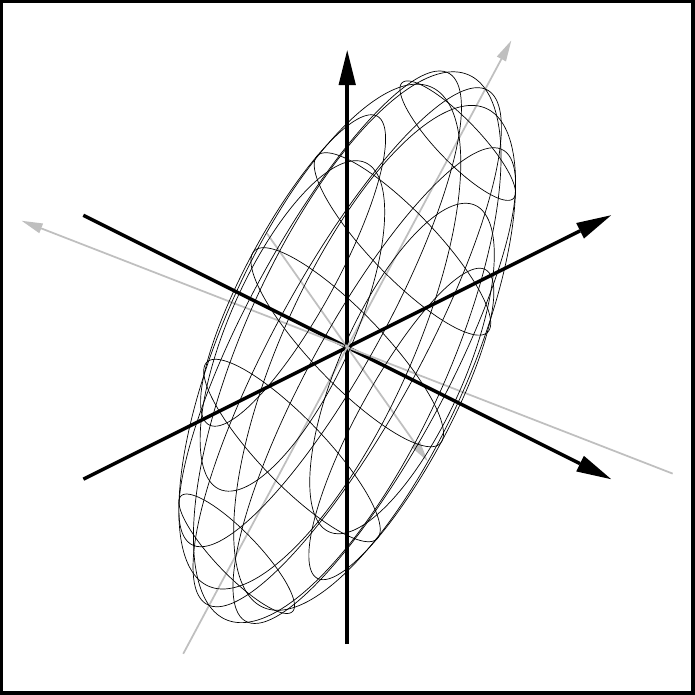}
	\caption{\label{figure:quadratic}
		A single level set of the functions sphere, ellipsoid, cigar,
		and discus. The first row shows separable variants of the
		functions. In the second row the functions are rotated (the
		rotated axes are displayed in gray), resulting in non-separable
		problems in all cases except for sphere, which is rotation
		invariant.
	}
\end{center}
\end{figure*}

Eigen-decomposition $H = UDU^T$ of the Hessian $H$ into an orthogonal
matrix $U$ of eigen vectors and a diagonal matrix $D$ holding the
(positive) eigen values is the central mathematical tool for describing,
understanding and \emph{constructing} this problem class. It can
represent two challenges: ill-conditioning and non-separability.
Ill-conditioning refers to a large conditioning number, defined as the
quotient $\kappa$ of maximal and minimal eigen values of $H$ (or $D$).
Separability plays a role only if the conditioning number is larger than
one. A separable problem is described by $U = \id$, which means that
there exist no cross-terms between variables, while a non-separable
problem is characterized by dependencies between variables, or
equivalently, by eigen vectors that are not axis-aligned. The sphere
function characterized by $H = U = D = \id$ is the simplest quadratic
problem. The ellipsoid problem features a spectrum of eigen values in
some range, often set to $\kappa = 10^6$, chosen equally spaced on a
logarithmic scale. Other benchmark problems of interest are the cigar
(single small eigen value, $d-1$ large eigen values) and discus (single
large eigen value, $d-1$ small eigen values) functions, usually with the
same $\kappa$ but different eigenvalue
distribution~\cite{hansen2010comparing}. The functions are named after
the shapes of their level sets, see figure~\ref{figure:quadratic} for an
illustration of the three-dimensional case.

Cholesky decomposition of the Hessian into $H = A A^T$ offers a
different perspective: the matrix $A$ can be thought of as a
transformation matrix between the ``intrinsic'' coordinate system of
the quadratic function (the coordinate system in which the Hessian
becomes the identity matrix) and the ``extrinsic'' coordinates in terms
of which the problem is stated \cite{wierstra:2014}. To make the
Cholesky factor unique it can be chosen either as a triangular matrix
with positive diagonal or as the symmetric positive definite matrix
$A = U \sqrt{D} U^T$.

\subsection{The Bi-Objective Case}

In this section we investigate convex quadratic bi-objective problems in
detail. There is surprisingly little literature on this topic. The
probably earliest quadratic bi-objective benchmark is Schaffer's pair of
one-dimensional spheres \cite{schaffer:1984}. Bi-objective BBOB
\cite{tusar2016coco} contains three purely quadratic functions in
arbitrary dimensions, namely combinations of sphere and separable
ellipsoid. The work of Augusto et al.\ \cite{augusto2014multiobjective}
focuses on two-dimensional problems and shows how to obtain the Pareto
front in the white-box case, see equation~\eqref{eq:set} below. A
similar analysis is found in \cite{kerschke2018search}. Here we are
interested in designing benchmarks for black-box solvers, but for goals
G3 to G6 an analytic understanding of the problem in the white-box sense
is crucial. A set of quadratic functions with different focus was
recently presented in~\cite{toure2019}.

Although convex quadratic functions are one of the simplest non-trivial
MO problem classes, we demonstrate that they pose significantly richer
challenges than in the single-objective case. Indeed, some of these were
apparently not considered in the existing literature.

Without loss of generality (see section~\ref{section:instances}) we
parameterize the objective functions as
\begin{align}
	f_i(x) &= \frac12 (x - x_i^*)^T H_i (x - x_i^*) \notag \\
	       &= \frac12 (x - x_i^*)^T A_i A_i^T (x - x_i^*) = \frac12 \big\| A_i^T (x - x_i^*) \big\|^2 \label{eq:objective} \\
	       &= \frac12 (x - x_i^*)^T U_i D_i U_i^T (x - x_i^*) \notag
\end{align}
for $i \in \{1, 2\}$, where $x_i^*$ is the optimum of $f_i$. The Hessian
matrices $H_i$ are symmetric and strictly positive definite. They are
represented by their eigen decomposition into orthogonal $U_i$ and
strictly positive definite diagonal $D_i$, or alternatively by their
Cholesky factors $A_i$. Furthermore we define $\delta = x_2^* - x_1^*$,
and in order to avoid trivial degenerate cases we assume
$\delta \not= 0$.

A point $x \in \R^d$ is Pareto optimal if for all $y \in \R^d$ it holds
either $f_1(x) \leq f_1(y)$ or $f_2(x) \leq f_2(y)$ (assuming
minimization of $f_1$ and $f_1$). For a smooth function a necessary
condition is that the gradients $\nabla f_i(x) = H_i (x - x_i^*)$ of the
two objectives cancel out
\cite{augusto2014multiobjective,kerschke2017expedition,toure2019},
i.e., that
\begin{align}
	c \cdot H_1 (x - x_1^*) + (1-c) \cdot H_2 (x - x_2^*) = 0
	\label{eq:geneig}
\end{align}
for some $c \in [0, 1]$. We obtain Pareto optimal solutions of the form
\begin{align}
	x = (c H_1 + (1-c) H_2)^{-1} \big[ c H_1 x_1^* + (1-c) H_2 x_2^* \big],
	\label{eq:set}
\end{align}
see also \cite{augusto2014multiobjective,toure2019}.
Hence, in a white-box setting the problem is solved analytically.
In accordance with goal G3 we want this set to be of a tractable form.
This is the case in particular if $\delta$ is a generalized eigenvector
of the pair $(H_1, H_2)$, i.e., if it holds
$g H_1 \delta - (1-g) H_2 \delta = 0$ for $g \in [0, 1]$.
Restricted to the affine line spanned by $x_1^*$ and $x_2^*$
equation~\eqref{eq:geneig} is a generalized eigen value problem. By
plugging the parameterization
$$
	x = (1-t) \cdot x_1^* + t \cdot x_2^* = x_1^* + t \delta = x_2^* - (1-t) \delta
$$
for $t \in [0, 1]$ into equation~\eqref{eq:geneig} we obtain
$$
	c t \cdot H_1 \delta = (1-c) (1-t) \cdot H_2 \delta.
$$
For each $t \in [0, 1]$ there exists $c \in [0, 1]$ such that it holds
$(1-c)(1-t)g = ct(1-g)$, so that the generalized eigen equation is
fulfilled. Therefore the Pareto set is given by the line segment
$$
	S = \Big\{(1-t) \cdot x_1^* + t \cdot x_2^* \,\Big|\, t \in [0,1]\Big\}.
$$
The corresponding Pareto front is convex \cite{toure2019}.
Let $n_1 = f_1(x_2^*)$ and $n_2 = f_2(x_1^*)$ denote the components of
the nadir point, then the Pareto front is the parameterized set
$$
	F = \left\{ \left. \left(t^2 \cdot n_1, \left(1 - t\right)^2 \cdot n_2 \right) \,\right|\, t \in [0, 1] \right\}.
$$

\subsection{Non-Separability and Ill-Conditioning}

%* Separability and ill-conditioning mean that the intrinsic coordinate
%system of problem and extrinsic coordinates are not well aligned. Here
%we have three coordinate systems, all of which can be misaligned.

As noted above, non-separability and ill-conditioning can be defined in
terms of the matrices $U_i$ and $D_i$. Here we take a slightly different
perspective based on the Cholesky factors $A_i$, which we think of as
transformations between the ``intrinsic'' coordinates in which the
quadratic problem is a sphere function and the given ``extrinsic''
coordinates. In the bi-objective case, three coordinate systems are
involved: the extrinsic system and two intrinsic ones. The standard
notion of non-separability and ill-conditioning asks how well or how
badly intrinsic systems are aligned with the extrinsic one, in the terms
defined above, which involve $U_i$ and $D_i$. However, we can equally
well ask whether the two intrinsic coordinate systems are aligned or
not. In the simplest case both objectives share the same Hessian, and in
the worst case their Hessians differ significantly. In the latter case,
the variations (e.g., mutations) that are most suitable for creating
successful offspring may vary when moving along the Pareto front.

Based on the above considerations we define a total number of nine cases
with different properties in terms of non-separability and
ill-conditioning:
\begin{itemize}
\item
	All three coordinate systems are aligned, i.e., $U_1 = U_2 = \id$.
	\begin{itemize}
	\item[(1)]
		Both functions are (shifted) spheres: $D_1 = D_2 = \id$ and
		hence $H_1 = H_2 = \id$.
	\item[(2)]
		One function is a sphere, the other one is not: $D_1 = \id \not= D_2$.
	\item[(3)]
		Both functions are ill-conditioned in the same way:
		$D_1 = D_2 \not= \id$.
	\item[(4)]
		Both functions are ill-conditioned in different ways:
		$\id \not= D_1 \not= D_2 \not= \id$.
	\end{itemize}
\item
	One intrinsic coordinate system is aligned with the extrinsic one,
	the other one is not: $U_1 = \id$, $U_2 \not= \id$, $D_2 \not= \id$.
	\begin{itemize}
	\item[(5)]
		$f_1$ is a sphere, i.e., $D_1 = \id$.
	\item[(6)]
		The Hessians are unrelated: $\id \not= D_2 \not= D_1 \not= \id$.
	\end{itemize}
\item
	No intrinsic coordinate system is aligned with the extrinsic one,
	but they are aligned with each other: $U_1 = U_2 \not= \id$ and
	$D_1 \not= \id \not= D_2$.
	\begin{itemize}
	\item[(7)]
		For $D_1 = D_2$ both problems share the same Hessian.
	\item[(8)]
		Otherwise the two functions are aligned but differently scaled.
	\end{itemize}
\item
	No pair of coordinate systems is aligned with each other:
	\begin{itemize}
	\item[(9)]
		$\id \not= U_1 \not= U_2 \not= \id$, $D_1 \not= \id \not= D_2$.
	\end{itemize}
\end{itemize}
Whenever matrices are unequal in the above list then, for the purpose of
sampling instances, we assume that they are ``significantly different''
with high probability. For orthogonal matrices this is the case for
example if $U_1^T U_2$ follows a uniform distribution on the set of
orthogonal transformations. For diagonal matrices we demand that
$D_1^{-1} D_2$ has a large conditioning number.

These classes fulfill all goals up to G6: They are scalable, we will
show in the next section how to create instances, the Pareto set is a
line segment in a special or general direction, the front is convex, for
which the optimal $\mu$-distribution is easy to obtain numerically with
the methods of \cite{auger:2009} or \cite{glasmachers2014b} (and this
needs to be done only once), and single-objective challenges like
non-separability and ill-conditioning are exactly what defines the
different problem classes. Goal G6 will be addressed in
section~\ref{section:G6}.

It is worth pointing out that the construction is more general than
rotating (essentially) separable benchmarks, as proposed in \cite{mocmaes}.
This is because we apply \emph{independent} affine transformations to
the two objectives, which provides great flexibility.

\subsection{Alignment of the Pareto Set}

The Pareto sets of all problems under consideration form line segments.
It makes sense to ask whether or not the line segment is aligned with
the extrinsic coordinate system. The answer depends on properties of the
Hessians since we always assume that $\delta$ is a generalized
eigenvector thereof. For some cases it is easy to construct aligned and
non-aligned instances, while in others we need to consider additional
constraints (see also section~\ref{section:construction}). At first
glance, cases 2, 3 and 4 demand that $\delta$ is axis-aligned since all
generalized eigen values have this property. However, even in this case
we can construct non-aligned Pareto sets. For this purpose we demand
that at least one generalized eigen space is at least two-dimensional.
Within this subspace the vector $\delta$ can be chosen freely. The
resulting vector is sparse (the number of non-zeros equals the dimension
of the eigen space), but not aligned with a single axis. This can be
sufficient to detect whether variation operators are able to move along
a non-aligned Pareto set.

\begin{figure*}
\begin{center}
	\includegraphics[width=0.32\textwidth]{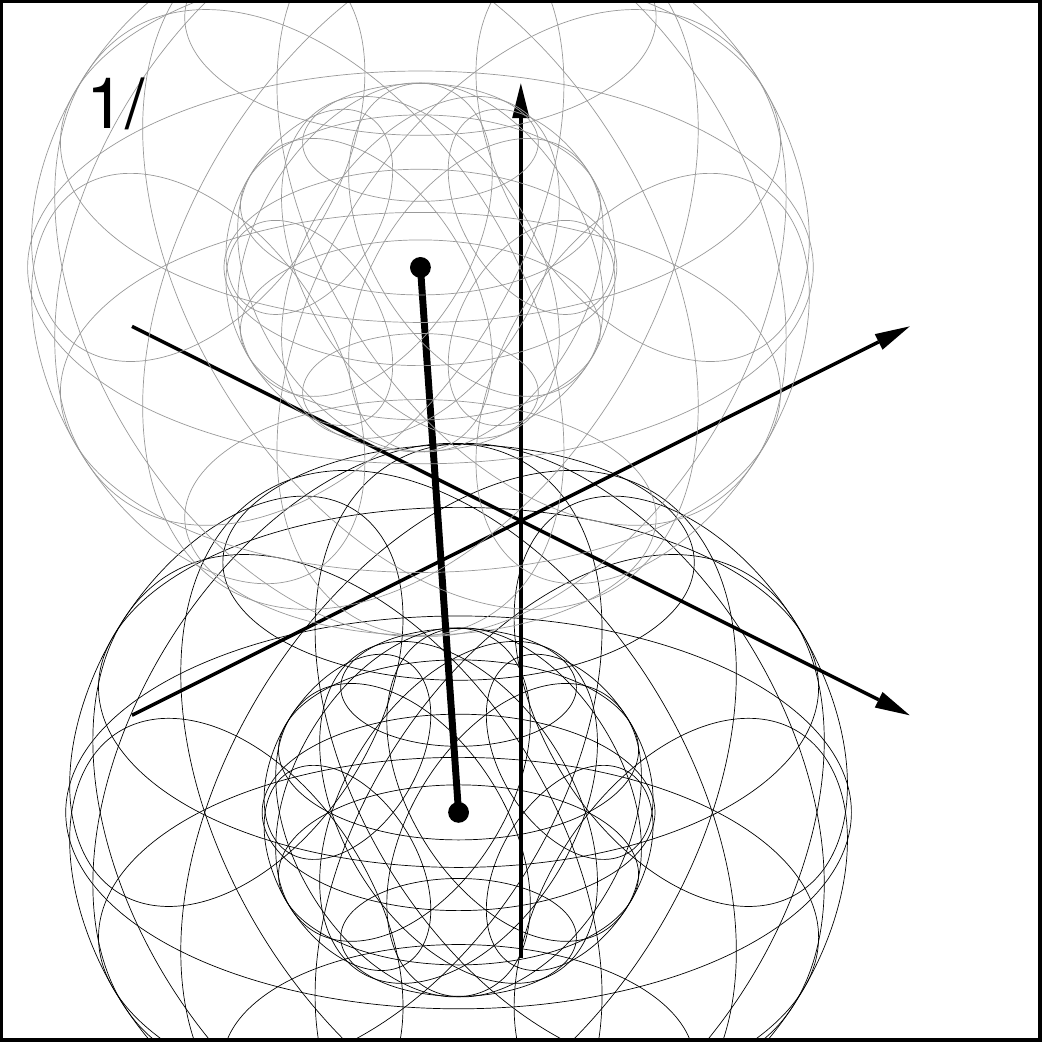}
	\includegraphics[width=0.32\textwidth]{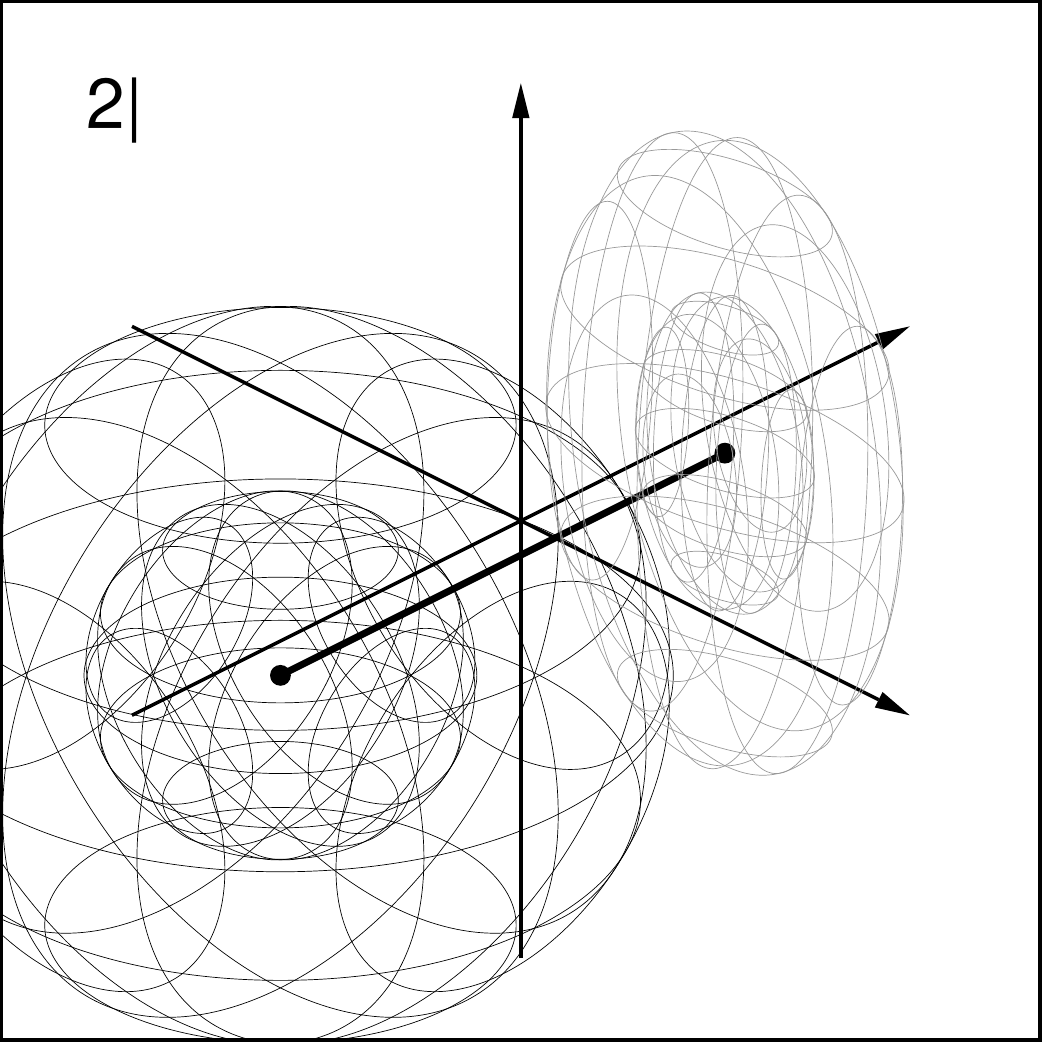}
	\includegraphics[width=0.32\textwidth]{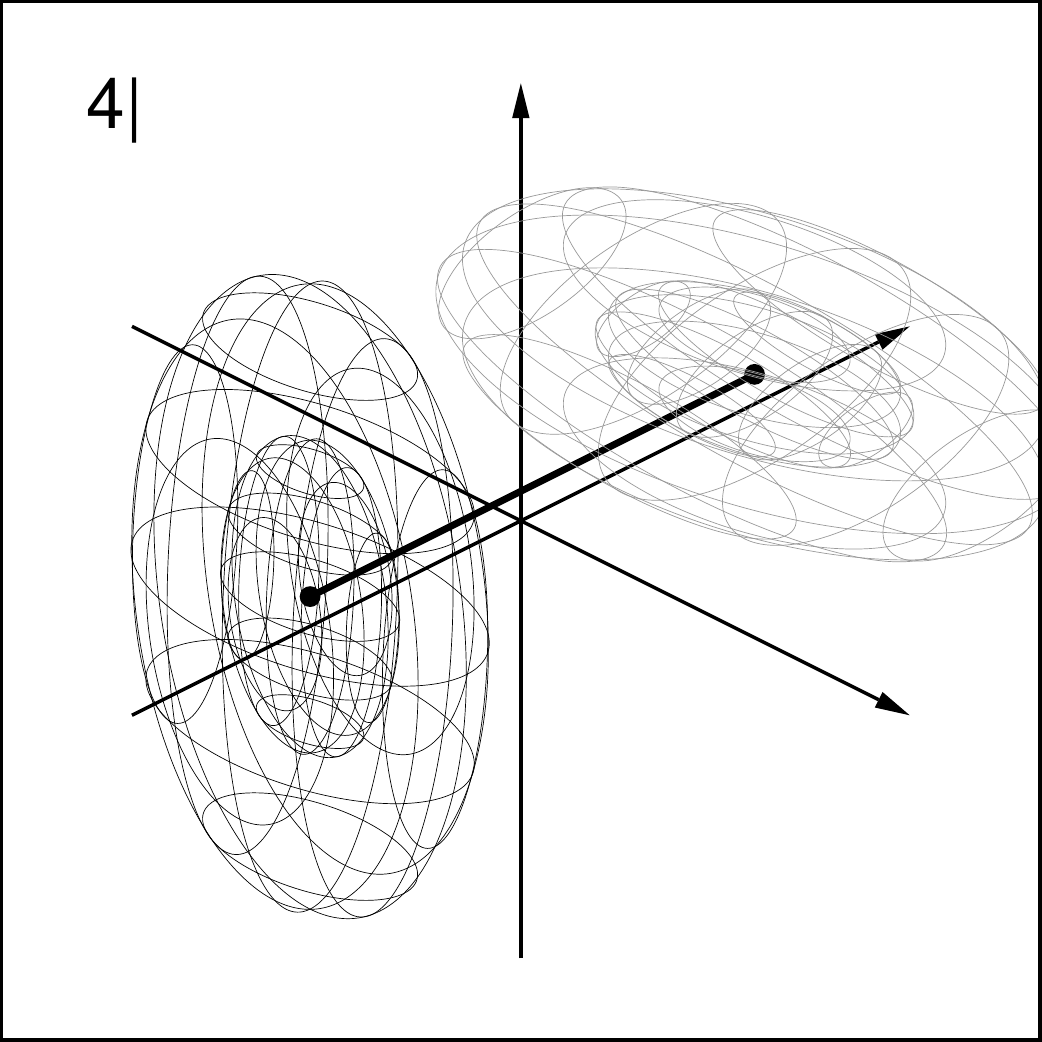}
	\\
	\includegraphics[width=0.32\textwidth]{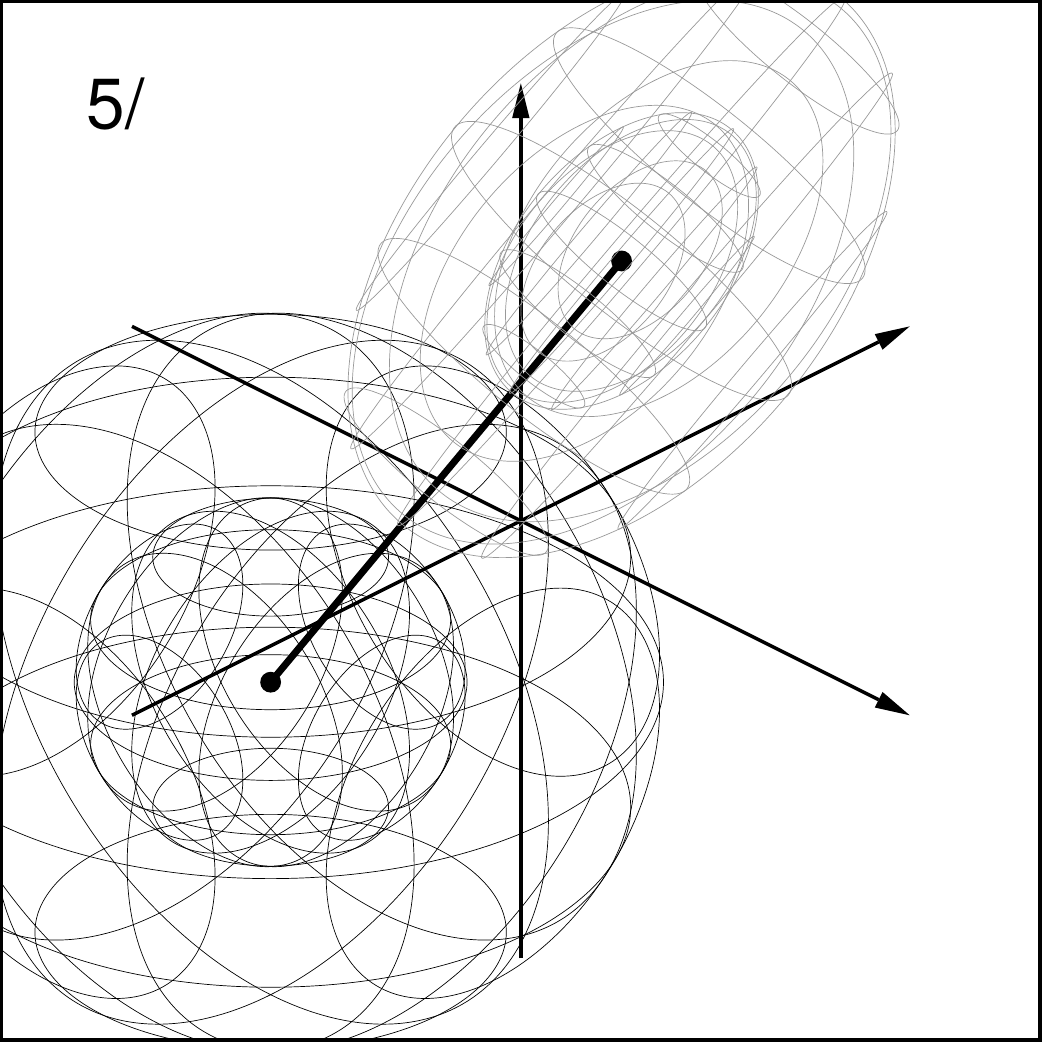}
	\includegraphics[width=0.32\textwidth]{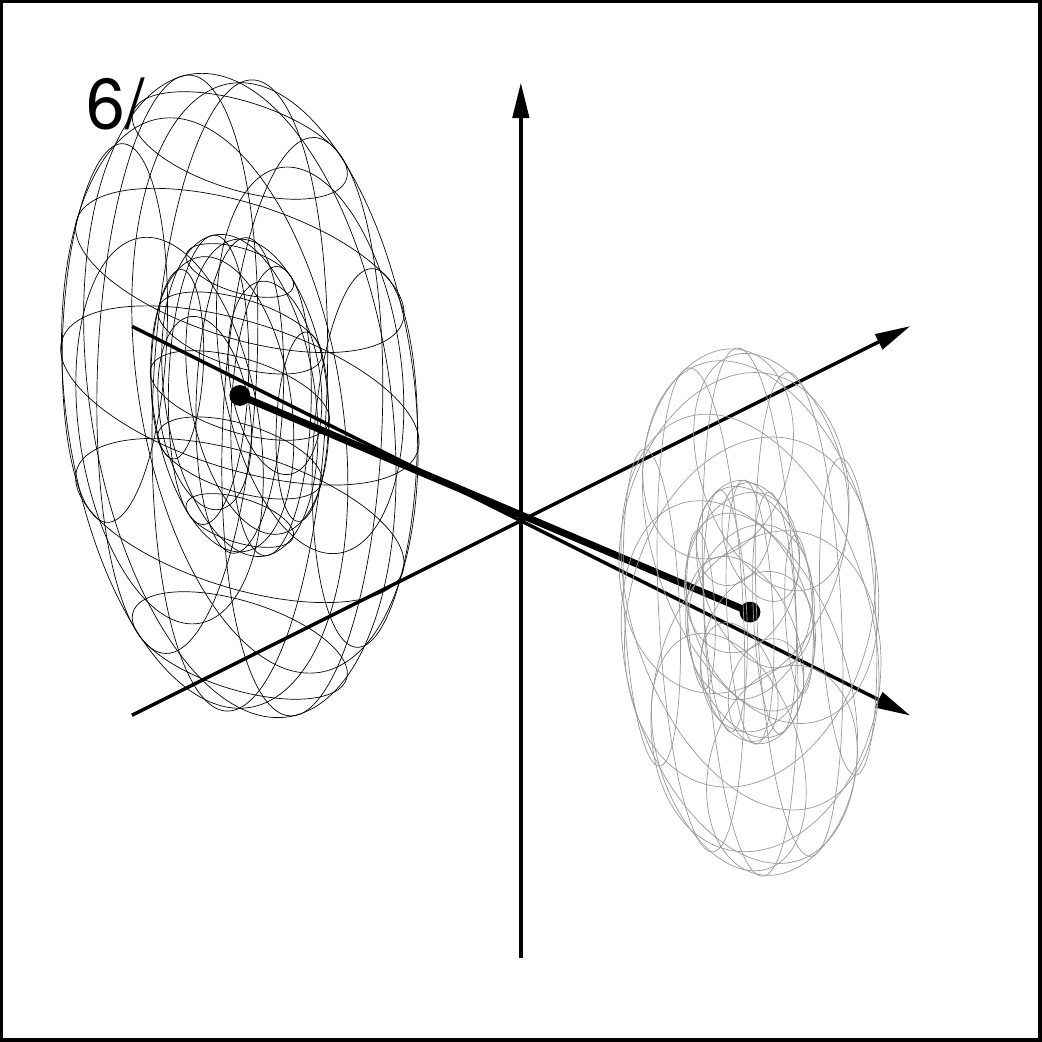}
	\includegraphics[width=0.32\textwidth]{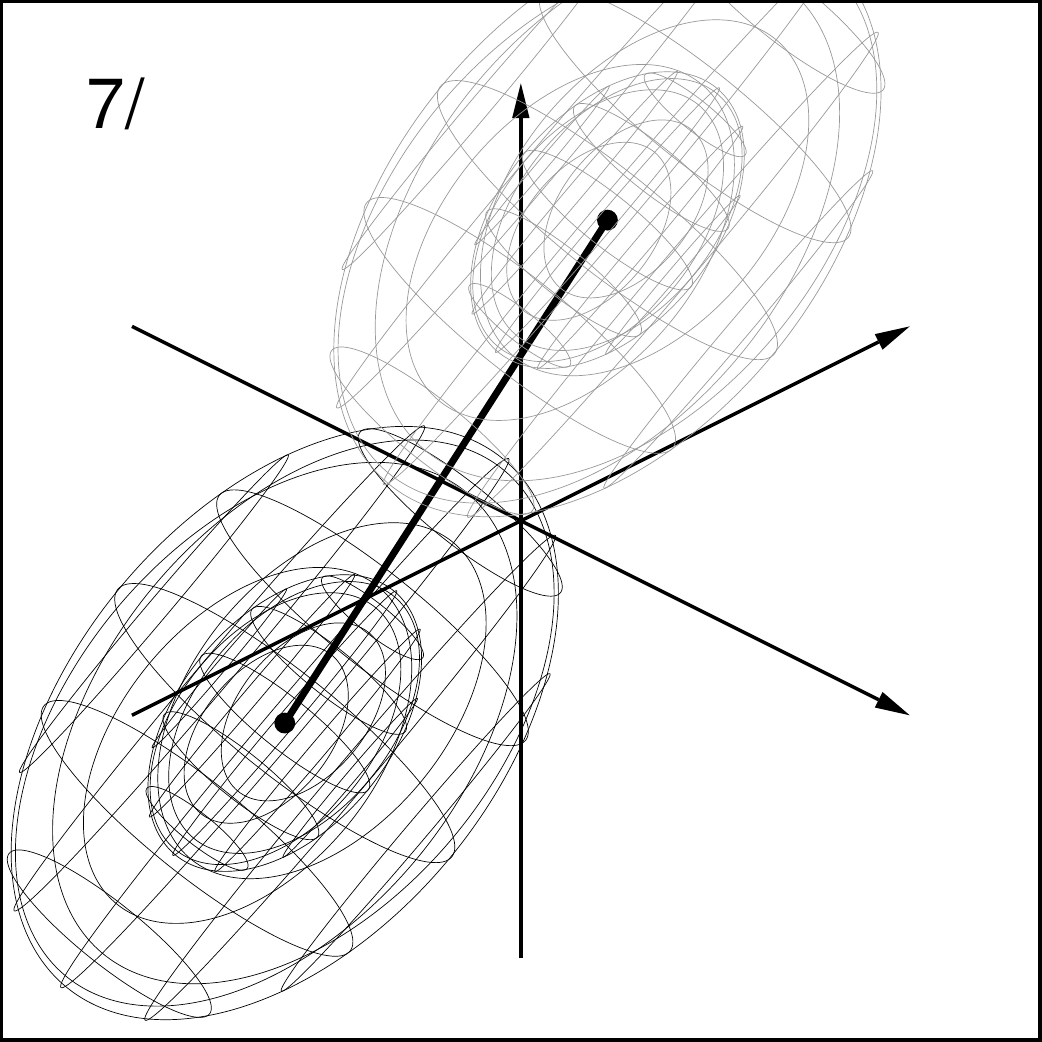}
	\\
	\includegraphics[width=0.32\textwidth]{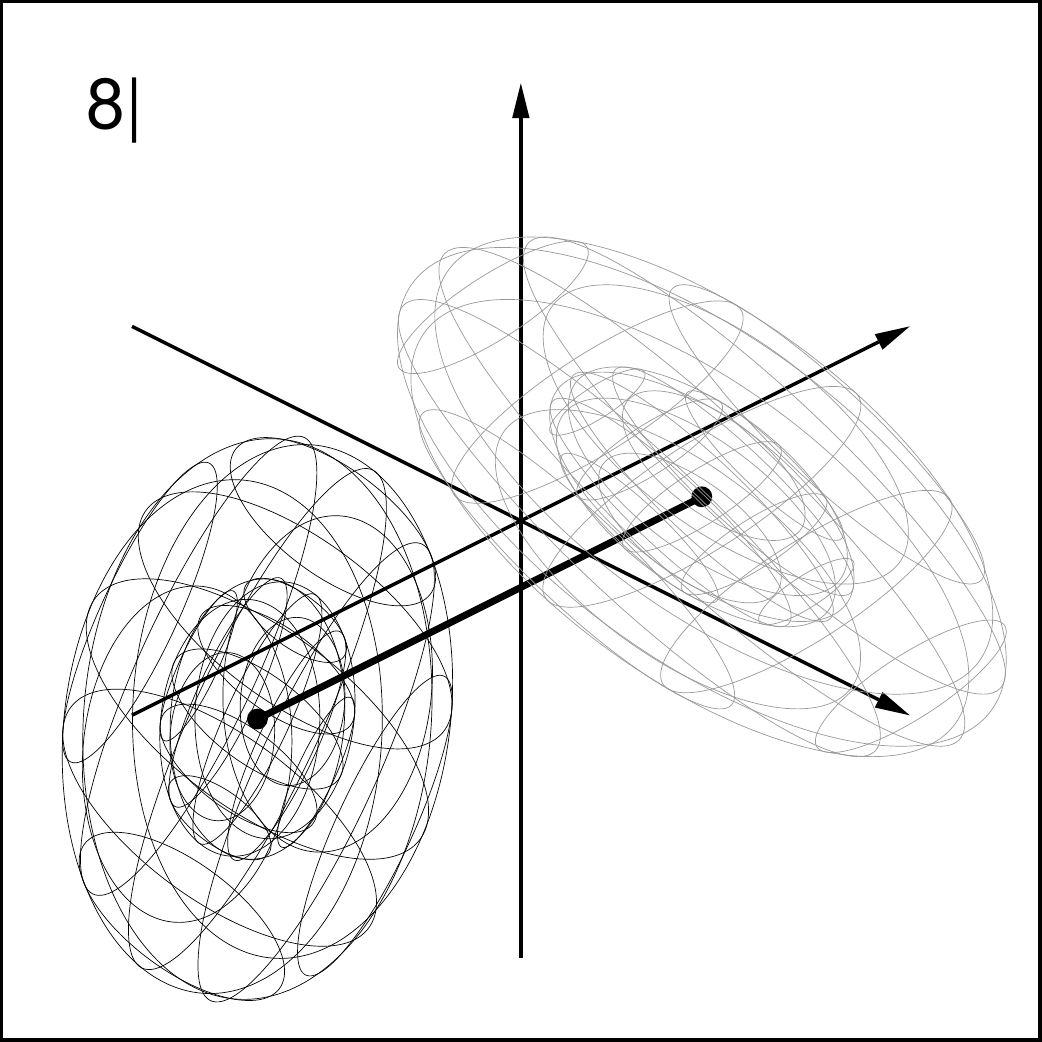}
	\includegraphics[width=0.32\textwidth]{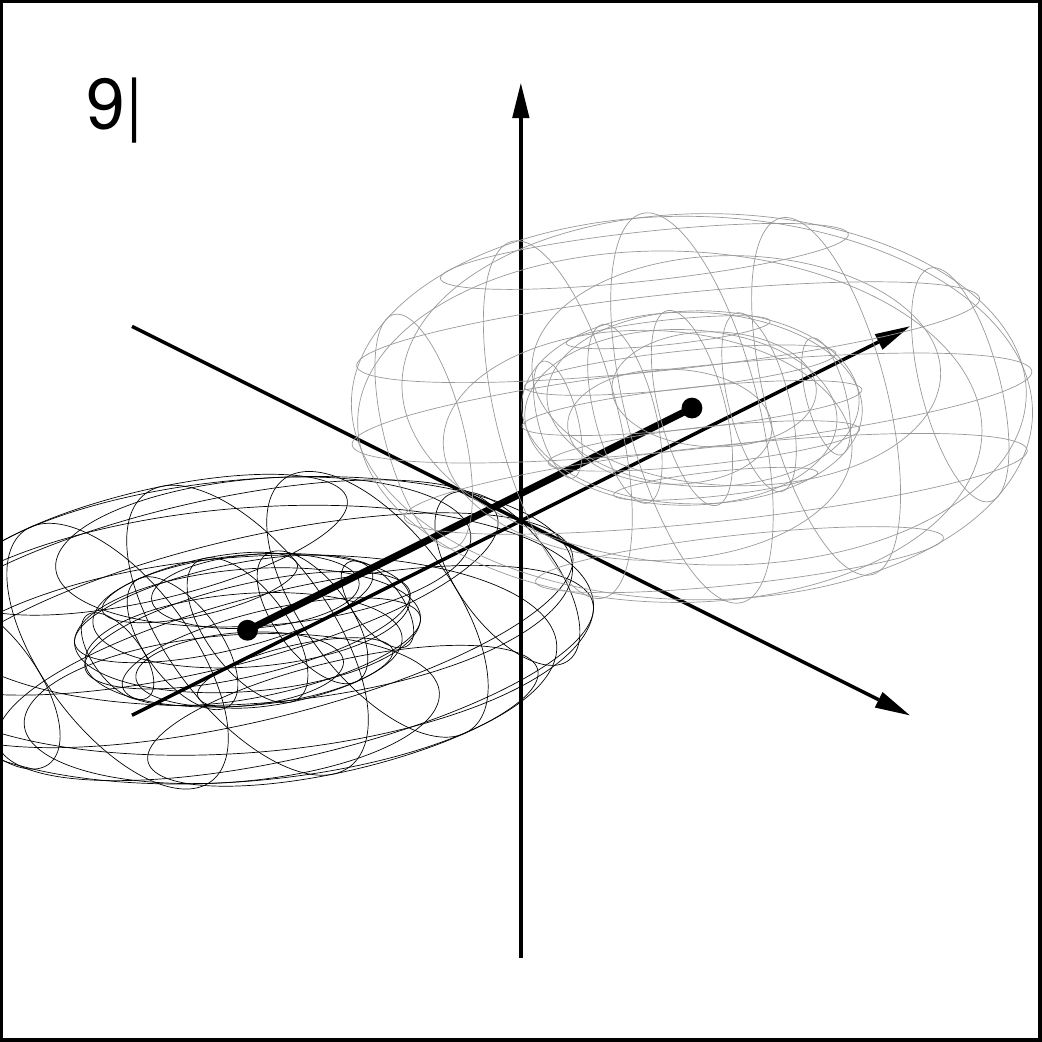}
	\includegraphics[width=0.32\textwidth]{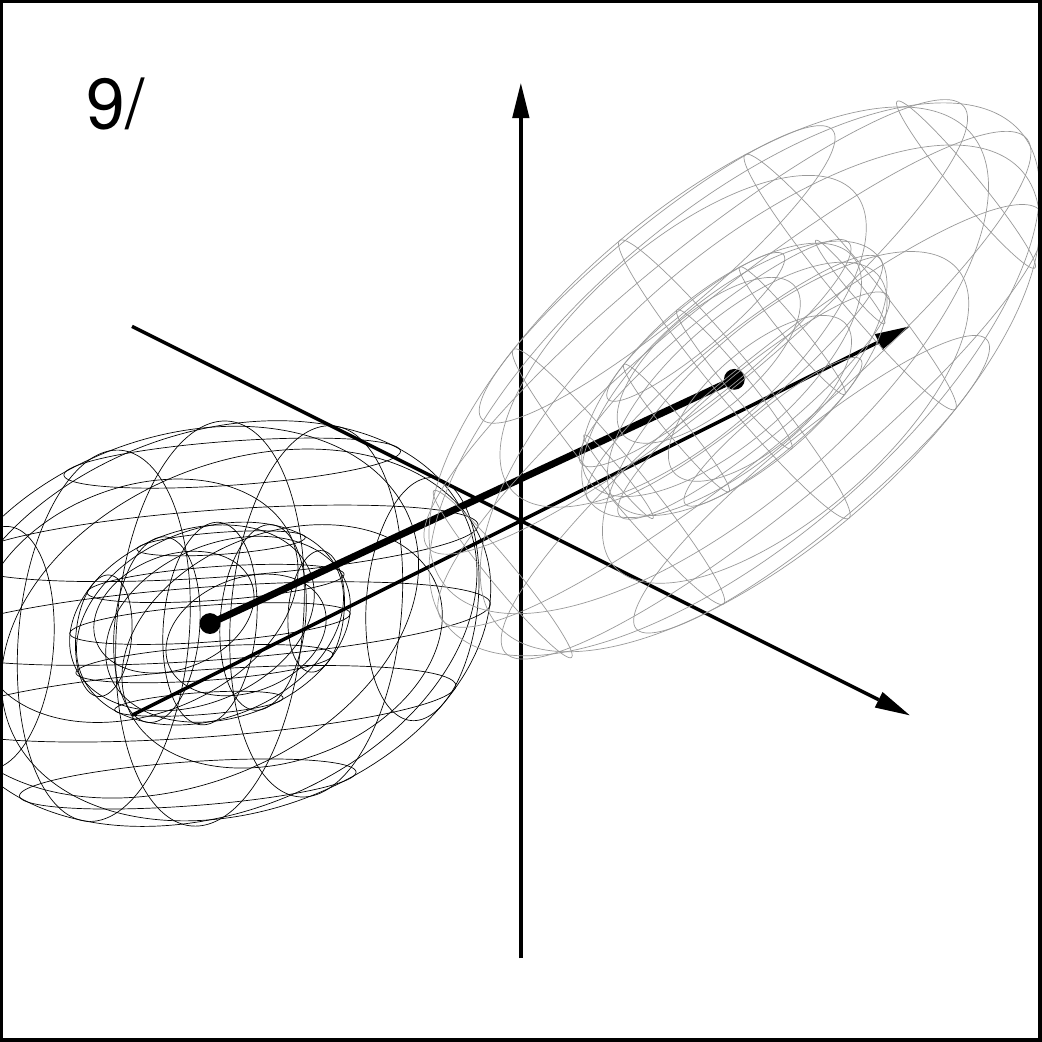}
	\caption{\label{figure:plots}
		Level sets of the two objective functions for 9 out of the 18
		quadratic problem classes, for a benign conditioning number of
		$\kappa = 10$. The black dots are the optima of the two
		objectives, and the line connecting the two is the Pareto set.
		The problems vary in level set shapes, level set alignment with
		each other and/or with the coordinate system, and alignment of
		the Pareto front with the coordinate system.
	}
\end{center}
\end{figure*}

With this construction, we extend the above scheme, consisting of nine
classes. For each class we consider two sub-classes, namely for
axis-aligned $\delta$ and for non-aligned $\delta$. The aligned case is
denoted by appending a vertical bar (ASCII pipe, \texttt{"|"}) to the
class number, and a tilted bar (ASCII slash, \texttt{"/"}) denotes the
non-aligned case. For example, \texttt{1/} is the class of problems
consisting of two sphere functions with optima $x_1^*$ and $x_2^*$ in
general position, while for class \texttt{7|} the objectives are jointly
rotated ellipsoid functions with axis-aligned Pareto set.
Two-dimensional example instances of some of the problem classes are
illustrated in figure~\ref{figure:plots}.

%\subsection{Movement Towards and Along the Front}
%
%%* Assume an ill-conditioned problem with the front aligned to one axis.
%%Then it should make a huge difference whether the axis is a short or a
%%long axis of the problem, i.e., whether it is easier to move towards the
%%front or along the front.
%
%Not only the conditioning number of a problem matters, but also which
%axes have low or high curvature (eigen value). Consider a simple
%two-dimensional case where one eigenvector of $H_1 = H_2$ is parallel to
%$\delta$. If the corresponding eigenvalue is small then it is relatively
%easy to generate successful points in this direction, which should help
%to spread out along the front quickly once it is found. However, if the
%eigenvalue is relatively large then moving towards the front is easier
%than spreading out. This could have an impact on which optimization
%strategies are successful: should one locate the front quickly and then
%spread out as proposed in \cite{glasmachers:2014}, or should one
%approach the front with many individuals in parallel?
%
%%% TODO: follow up on this question in the experiments, and it should
%%%       actually also be reflected in the construction; otherwise just
%%%       remove the sub-section

\subsection{Multi-Objective Challenges}
\label{section:G6}

In this section we consider challenges that are specific to
multi-objective optimization. This is a wide field, and in order not to
deviate too widely from quadratic function, here we only scratch the
surface.

As shown in equation~\eqref{eq:objective}, every quadratic function
$f(x)$ can
be written as the sphere function $g(x) = \frac12 \|x\|^2$ applied to
the linearly transformed point $h(x) = A^T (x - x^*)$:
$f(x) = g\big(h(x)\big)$. The whole power of the construction until now
relies on flexibly constructed transformations~$h$. Multi-objective
challenges are best constructed by varying~$g$.

A simple modification that does not even change the level sets of the
individual objectives is to use a different power for the Euclidean norm:
$g(x) = \frac12 \|x\|^s$. A pair of quadratic spheres ($s=2$) yields a
convex front. A linear front is obtained for $s=1$, and a concave front
for example with $s=1/2$ (a quarter of an ellipse)
\cite{emmerich2007test}. We mark these cases by appending the upper case
letters \texttt{C}, \texttt{I} and \texttt{J} to the problem name,
resembling a convex, linear, and concave shape.

Similar constructions were applied in the construction of the ZDT, DTLZ,
and WFG functions. These function classes go beyond our approach in
several respects, and they provide systematic construction kits. In
order to limit the complexity of our construction we do not follow this
well-explored road, and we limit goal~G6 to the above three shapes of
the Pareto front. Anyway, it is worth mentioning that it should be
possible to combine our construction with other challenges by applying
the 18 different classes of transformations first, followed by existing
techniques, which are suitable for generating disconnected, partially
flat, combined convex/concave, or otherwise interesting front shapes.

\subsection{Extension to $m > 2$ Objectives}

It is straightforward to extend our construction to more than two
objectives. However, making sure that the Pareto set is of a tractable
form is a bit more challenging, since it requires pairwise compatibility
of all Hessians. Also, a fully factorial design would result in a
combinatorial explosion of different cases. This is not only
undesirable, but also unnecessary. For example, it is probably
sufficient to test cases where subsets of the objective functions use an
aligned coordinate system. Instead, either all functions should either
be jointly or independently rotated, or none. Also, an $m-1$ dimensional
Pareto set can be fully aligned to the coordinate system or be in
general position, while partial alignment is unlikely to provide
additional insights. This way it is possible to stick to the 18 classes
of transformations identified above.

\section{Benchmark Construction}
\label{section:construction}

In this section we discuss how to construct instances of the benchmark
problem classes. To this end we will fix a few remaining design
decisions. The main problem tackled in this section is how to sample
from spaces of function under a variety of different constraints
defining the different problem classes.

In the following we will always assume that $\delta$ is a generalized
eigenvector of $(H_1, H_2)$. Technically this can be achieved by setting
$\delta$ first and by constructing appropriate matrices $H_i$, or by
fixing $H_i$ arbitrarily and picking $\delta$ as one of the generalized
eigen vectors. We will use both approaches in the following, depending
on the constraints posed by the different cases.

We aim to construct all 18 classes of transformations according to the
same scheme, as far as possible. We first discuss a straightforward
approach that works only in the absence of constraints, and then turn to
sampling procedures for constrained cases.

\subsection{Sampling with and without Constraints}

If an orthogonal matrix $U_i$ differs from the identity then we sample
it from the uniform distribution on the orthogonal matrices. An easy to
implement sampling procedure is to create a $d \times d$ matrix with
entries sampled i.i.d.\ from the standard normal distribution
$\Normal(0, 1)$, and to apply the Gram-Schmidt orthogonalization
process. For diagonal matrices that differ from the identity we apply
the eigen value spectrum of the ellipsoid function
$D_{ii} = \kappa^{\frac{i-1}{n-1}}$ with default value $\kappa = 10^3$,
corresponding to a non-trivial yet moderate conditioning number.
However, we randomly permute the eigen values. The point $(x_1^* +
x_2^*)/2$ is sampled from the multivariate standard normal distribution
$\Normal(0, \id)$ under the additional constraint that no component
exceeds the range $[-4.5, 4.5]$. The vector $\delta$ is defined as a
generalized Eigen vector of the pair $(H_1, H_2)$ selected uniformly at
random and scaled to unit length. This uniquely determines the
points~$x_i^*$. The construction guarantees that the Pareto set is
contained in the hypercube $[-5, 5]^d$. The ability to provide bounds
makes variation operators for bounded spaces applicable as they are used
by several popular algorithms.

The above procedure is unable to produce instances of cases \texttt{2/},
\texttt{3/}, and \texttt{4/}, which require an at least two-dimensional
generalized eigen space. We therefore use the eigen values of the
$(d-1)$-dimensional case and duplicate one eigen value at random.
Furthermore we ensure that the duplicate eigen values are in the same
position in both diagonal matrices. Then $\delta$ is non-zero only in
these two coordinates (called $i$ and $j$), with
$\delta_i = \cos(\alpha)$ and $\delta_j = \sin(\alpha)$ with $\alpha$
uniform on $[0, 2\pi]$.

Another constraint is posed in cases \texttt{5|}, \texttt{6|},
\texttt{7|}, \texttt{8|}, \texttt{9|}, which require an axis-aligned
vector $\delta$ despite non-trivial orthogonal transformations being in
place. In cases \texttt{5|} and \texttt{6|} this is achieved by
modifying the random matrices before applying Gram-Schmidt
orthogonalization. We fix a random index $i \in \{1, \dots, d\}$ and set
the $i$-th row and column to zero, while the diagonal value is set to
one. In cases \texttt{7|} and \texttt{8|} the same procedure is applied
to both random matrices with the same index~$i$. In these cases,
$\delta$ is the vector consisting of $d-1$ zeros with a single one in
position~$i$.

Case \texttt{9|} is solved differently. First all matrices are sampled
as in case \texttt{9/}, and $\delta$ is determined. Then a further
orthogonal matrix $U$ is created based on a matrix with i.i.d.\ Gaussian
entries, however, with the first column fixed to $\delta$. When
orthogonalized, the Gram-Schmidt process starts with the first column of
the matrix and hence leaves $\delta$ unchanged. After orthogonalization,
$\delta$ is swapped with a random column. Finally, $U_1$ and $U_2$ are
replaced with $U^T U_1$ and $U^T U_2$, and we replace $\delta$ with $U^T
\delta$.

\subsection{Reproducible Instances}
\label{section:instances}

Instances of the above defined problem classes can be constructed by
following the sampling procedures defined above. The easiest way to
define reproducible instances is to fix the seed of the random number
generator used for sampling. This yields a nearly arbitrarily large
number of instances at no additional cost. We use a 19937-bit Mersenne
twister for this purpose.

Our sampling procedure does already produce instances that vary in terms
of typical transformations of the input space like translation, rotation
(as far as possible within the problem class), and permutation of
variables. However, all functions are still defined on the same scale of
values, and the componentwise optima have a value of zero. In order to
avoid any bias in the evaluation we also vary these.

To this end we argue that strictly monotone transformations of the
objective functions do not change the Pareto set \cite{toure2019} and
the level sets of the component functions, and hence can be considered
simple variants of the same problem class. We propose to apply affine
linear transformations of the form $a_i f_i + b_i$ with $a_i > 0$. We
sample $\log_{10}(a_i)$ and $b_i$ from the uniform distributions on
$[0, 6]$ and $[-a_i, +a_i]$, respectively.

At this point our construction has the general form
\begin{align*}
	f_i(x) = \frac{a_i}{2} \left[(x - x_i^*)^T U_i D_i U_i^T (x - x_i^*)\right]^{s/2} + b_i.
\end{align*}
It represents a total of $9 \cdot 2 \cdot 3 = 54$ function classes. This
full factorial design allows for a detailed analysis of factors
impacting performance.

A problem instance is defined by the parameters $a_i$, $b_i$, $U_i$,
$D_i$, $x_i^*$, and $s$. These parameters can be obtained from the
sampling procedures defined above for all 18 classes of transformations
with their different constraints.

\section{Experiments}

We have implemented our benchmark collection in C++ based on the Shark
machine learning library \cite{shark:2008}. Given a problem class name,
a dimension and an instance index it creates a problem instance that can
be evaluated. The code is provided in the supplementary material.

In order to exemplify the insights that can be gained just from
quadratic benchmarks we run experiments with
MO-CMA-ES \cite{mocmaes} with hypervolume-based selection, SMS-EMOA
\cite{smsemoa}, and NSGA-II \cite{nsga2}. We also tried NSGA-III, but it
performed consistently worse than its predecessor, simply because the
newer method is designed for many-objective problems, while we consider
only $m=2$ objectives. All methods were applied with their default
operators as implemented in the Shark library~\cite{shark:2008}.

Let $u$ and $n$ denote utopian and nadir point, respectively. We define
the reference point for the calculation of the dominated hypervolume as
$\frac{11n-u}{10}$, which amounts to shifting the nadir by $10\%$ of the
distance between utopian point and nadir point. This setting ensures
that identifying the extreme points pays off \cite{auger:2009}, but
without over-emphasizing their role. The hypervolume itself differs by
many orders of magnitude, depending on the problem instance. In order to
obtain a less problem dependent measure, we divide it by the area
between utopian point and nadir point, which is of size
$(n_1 - u_1) \times (n_2 - u_2)$. We refer to the quotient as the
normalized dominated hypervolume.

\begin{figure*}
\begin{center}
	\includegraphics[width=0.49\textwidth]{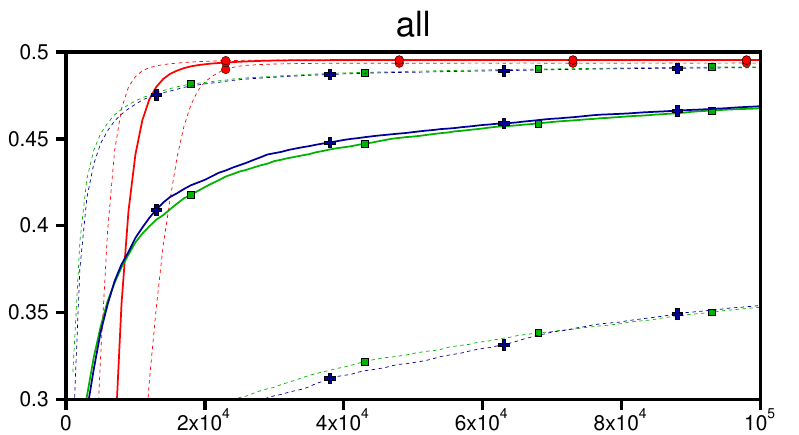}
	\includegraphics[width=0.49\textwidth]{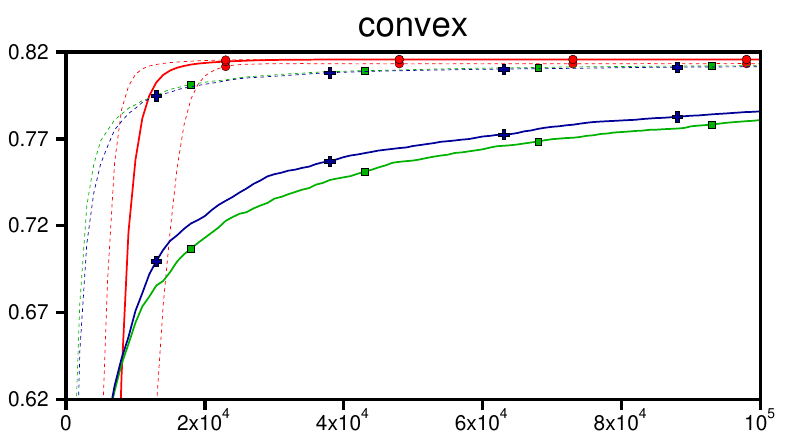}
	\\[2mm]
	\includegraphics[width=0.49\textwidth]{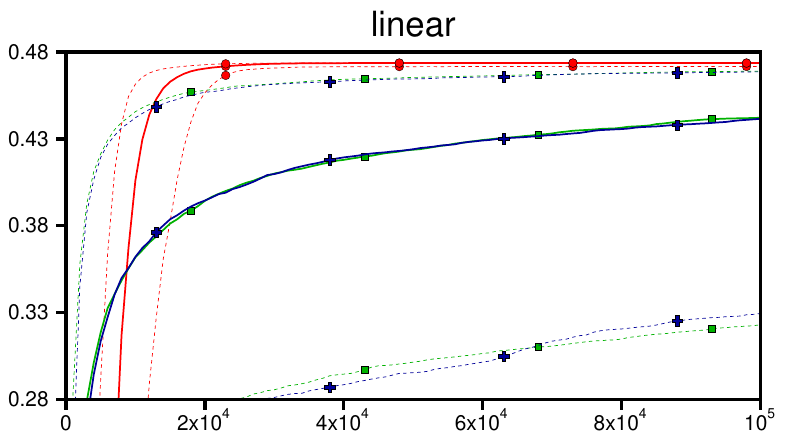}
	\includegraphics[width=0.49\textwidth]{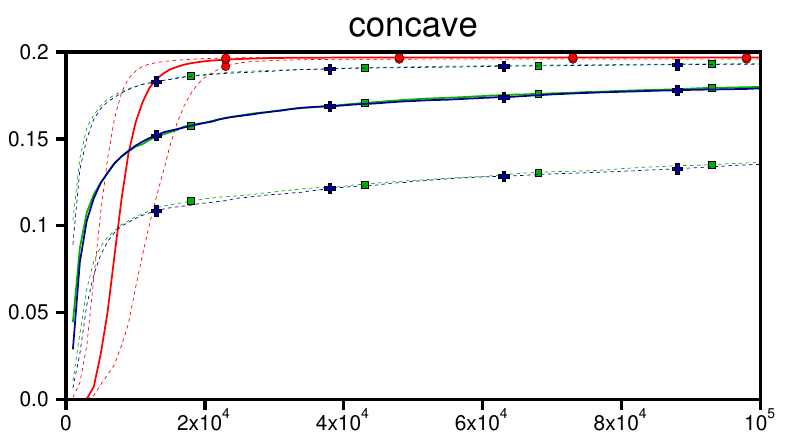}
	\\[2mm]
	\includegraphics[width=0.49\textwidth]{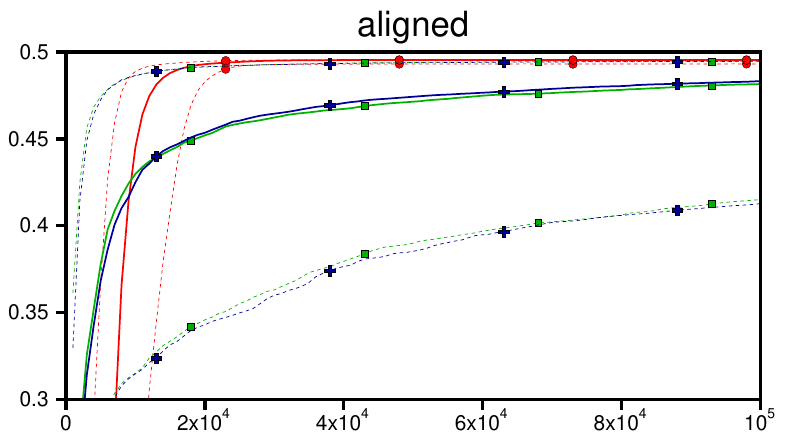}
	\includegraphics[width=0.49\textwidth]{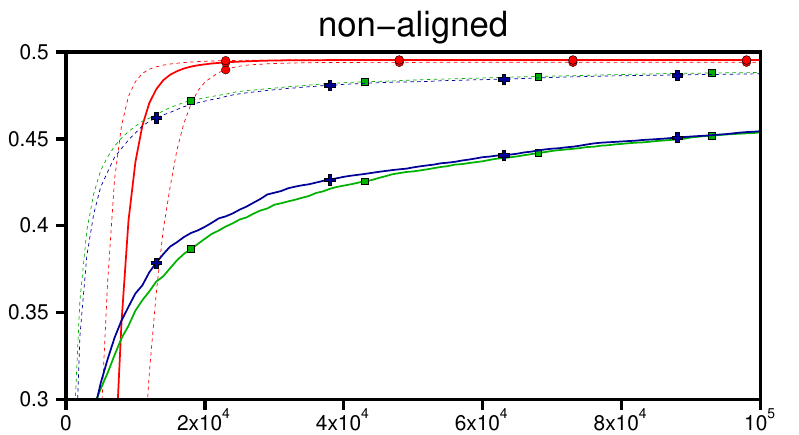}
	\\[2mm]
	\includegraphics[width=0.49\textwidth]{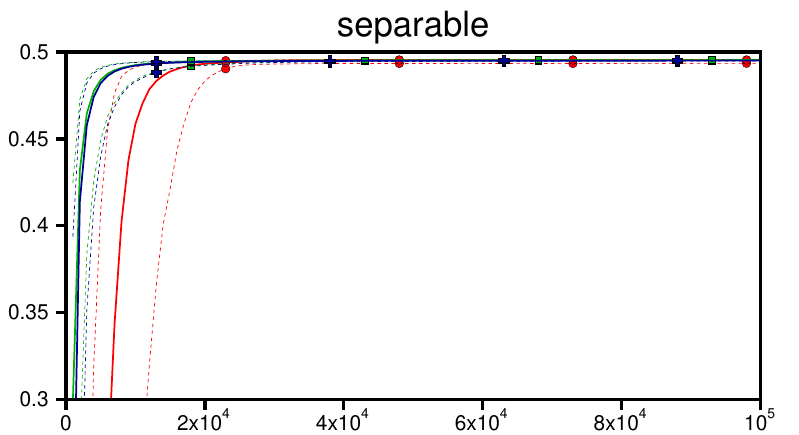}
	\includegraphics[width=0.49\textwidth]{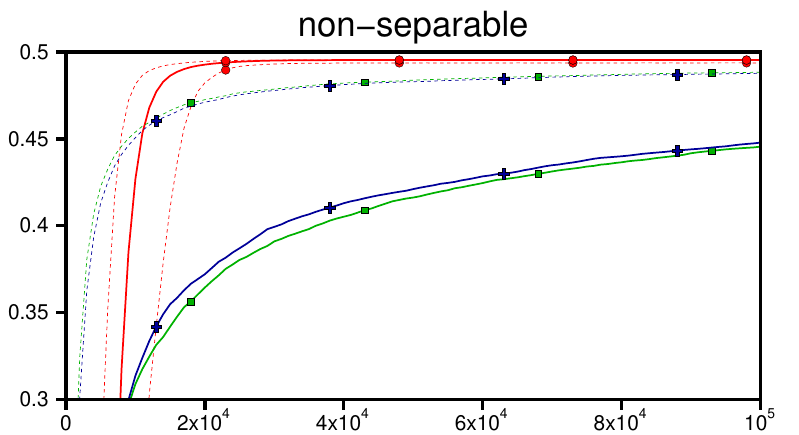}
	\\[3mm]
	\includegraphics[width=0.98\textwidth]{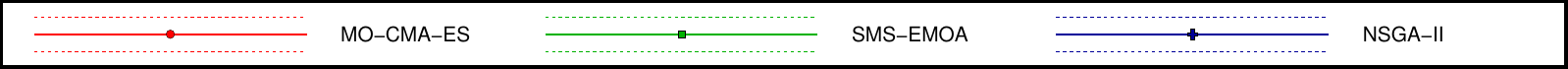}
	\caption{\label{figure:aggregates}
		Average performance of MO-CMA-ES, SMS-EMOA, and NSGA-II over
		eight different problem groups. The plots show dominated
		hypervolume, averaged over problem classes and 101 instances per
		problem, over number of fitness evaluations. The solid curves
		are medians over 101 problem instances, the dashed curves are
		the corresponding 10\% and 90\% quantiles.
	}
\end{center}
\end{figure*}

We ran the three above-mentioned algorithms on 101 instances of each of
the 54 problem classes defined in the previous section. The search space
was of dimension $d = 10$, and we set the population size of all three
algorithms to $\mu = 20$. The bounds of the variables for SMS-EMOA and
NSGA-II were set to $[-5, 5]$ (containing the Pareto set), and
MO-CMA-ES was initialized at the origin with an initial step size of
$\sigma = 3$, yielding a roughly comparable initial distribution. Each
solver was given a (rather generous) budget of $100,000$ fitness
evaluations.

This setup is by no means a distinguished one, and one could experiment
with many parameters, in particular with the problem dimension $d$ and
with the population size~$\mu$. Here we refrain from an extensive
experimental study, since in this work comparing different algorithm is
only a side product, while our central aim is to demonstrate our class
of convex quadratic multi-objective benchmarks.

\subsection{Results}

The evolution of the normalized dominated hypervolume is shown in
figure~\ref{figure:aggregates} for eight selected groups of functions.
Of course, depending on the research question each function can be
considered individually, and it can be contrasted with other functions
to demonstrate specific effects. Here we focus on groups identified
already in section two, namely different shapes of the front,
axis-aligned and non-aligned Pareto sets, and separable vs.\
non-separable problems.

\subsection{Discussion}

Figure~\ref{figure:aggregates} reveals tremendous differences in
performance. Generally speaking, MO-CMA-ES excels on all problems, and
its performance is nearly constant across all problems groups. This is
no surprise, given its very good performance on quadratic
single-objective problems, for which covariance matrix adaptation is a
perfect match, as well as its invariance properties.

In contrast, the variation operators of SMS-EMOA and NSGA-II are
beneficial only in an early phase, and later on for separable problems.
The inability of the variation operators to model directions other than
the given problem axes apparently causes a severe degradation in
performance in all other cases.

In contrast, the shape of the front has a minor impact on performance
(note that the achievable dominated hypervolume differs, while all plots
show a range of 0.2, measured in normalized dominated hypervolume).

The question whether the Pareto set is aligned with a problem axis or
not has a larger impact, and a more fine-grained analysis (analyzing
plots for each single transformation, not shown for space limitations)
reveals that the difference is most pronounced for cases \texttt{8|} and
\texttt{8/} and \texttt{9|} and \texttt{9/}, where none of the objective
functions is aligned with the coordinate system.

An even more pronounced degradation of optimization performance for
SMS-EMOA and NSGA-II is caused by rotating the coordinate system, which
causes non-separability.

In summary, we can conclude that the variation operators of NSGA-II and
SMS-EMOA are not well suited as soon as the Pareto set is not aligned
with a coordinate axis, and as soon as the problem is non-separable.
These seem to be very reasonable assumptions for black-box real-world
problems. Of course, due to the very nature of convex quadratic
functions, we can conclude superiority of MO-CMA-ES only for the
challenges of non-separability and ill-conditioning, which are by no
means the only challenges in multi-objective optimization problems.

\section{Conclusion}

We have constructed an expressive set of problem classes for testing
bi-objective optimization algorithms from rather simple primitives,
namely from convex quadratic functions. Our construction procedure
addresses shortcomings of existing benchmark collections, and we hope
that the 18 transformation classes will enter future benchmark
construction procedures.

We discovered novel challenges that were not considered before (to the
best of our knowledge), like alignment of the function with the given
coordinate system (separability) vs.\ alignment of the objectives with
each other, and the distinction between separability and the alignment
of the Pareto set with the coordinate axes.

Despite the limited versatility of convex quadratic functions, our
benchmark collection exhibits high discriminative power as demonstrated
in the experiments, where extreme performance differences can be
observed and clearly attributed to properties of the benchmarks, which
are known by construction. We have not yet exploited the full potential
of the construction, which also provides full knowledge of Pareto set
and Pareto front. Such information is invaluable for algorithm
comparison and analysis purposes.

\end{multicols}

\end{document}